# Dependency Networks for Collaborative Filtering and Data Visualization

**David Heckerman, David Maxwell Chickering, Christopher Meek, Robert Rounthwaite, Carl Kadie**
Microsoft Research
Redmond WA 98052-6399
heckerma,dmax,meek,robertro,carlk@microsoft.com

## Abstract

We describe a graphical representation of probabilistic relationships—an alternative to the Bayesian network—called a dependency network. Like a Bayesian network, a dependency network has a graph and a probability component. The graph component is a (cyclic) directed graph such that a node's parents render that node independent of all other nodes in the network. The probability component consists of the probability of a node given its parents for each node (as in a Bayesian network). We identify several basic properties of this representation, and describe its use in collaborative filtering (the task of predicting preferences) and the visualization of predictive relationships.

## 1 Introduction

The Bayesian network has proven to be a valuable tool for encoding, learning, and reasoning about probabilistic relationships. In this paper, we introduce another graphical representation of such relationships called a *dependency network*. The representation can be thought of as a collection of regression/classification models among variables in a domain that can be combined using Gibbs sampling to define a joint distribution for that domain. The dependency network has several advantages and disadvantages with respect to the Bayesian network. For example, a dependency network is not useful for encoding causal relationships and is difficult to construct using a knowledge-based approach. Nonetheless, in our three years of experience with this representation, we have found it to be easy to learn from data and quite useful for encoding and displaying predictive (i.e., dependence and independence) relationships. In addition, we have empirically verified that dependency networks are well suited to the task of predicting preferences—a task often referred to as *collaborative filtering*. Finally, the representation shows promise for density estimation and probabilistic inference.

The representation was conceived independently by Hofmann and Tresp (1997), who used it for density estimation; and Hofmann (2000) investigated several of its theoretical properties. In this paper, we summarize their work, further investigate theoretical properties of the representation, and examine its use for collaborative filtering and data visualization.

In Section 2, we define the representation and describe several of its basic properties. In Section 3, we describe algorithms for learning a dependency network from data, concentrating on the case where the local distributions of a dependency network (similar to the local distributions of a Bayesian network) are encoded using decision trees. In Section 4, we describe the task of collaborative filtering and present an empirical study showing that dependency networks are almost as accurate as and computationally more attractive than Bayesian networks on this task. Finally, in Section 5, we show how dependency networks are ideally suited to the task of visualizing predictive relationships learned from data.

## 2 Dependency Networks

To describe dependency networks and how we learn them, we need some notation. We denote a variable by a capitalized token (e.g., $X, X_i, \Theta$, Age), and the state or value of a corresponding variable by that same token in lower case (e.g., $x, x_i, \theta$, age). We denote a set of variables by a bold-face capitalized token (e.g., $\mathbf{X}, \mathbf{X}_i, \mathbf{Pa}_i$). We use a corresponding bold-face lowercase token (e.g., $\mathbf{x}, \mathbf{x}_i, \mathbf{pa}_i$) to denote an assignment of state or value to each variable in a given set. We use $p(X = x|Y = y)$ (or $p(x|y)$ as a shorthand) to denote the probability that $X = x$ given $Y = y$. We also use $p(x|y)$ to denote the probability distribution for $X$ given $Y$ (both mass functions and density functions). Whether $p(x|y)$ refers to a probability, a probability density, or a probability distribution will be clear from context.

Consider a domain of interest having variables $\mathbf{X} = (X_1, \ldots, X_n)$. A dependency network for $\mathbf{X}$ is a pair $(\mathcal{G}, \mathcal{P})$ where $\mathcal{G}$ is a (cyclic) directed graph and $\mathcal{P}$ is a set of probability distributions. Each node in $\mathcal{G}$ corresponds to a variable in $\mathbf{X}$. We use $X_i$ to refer to both the variable and its corresponding node. The parents of node $X_i$, denoted $\mathbf{Pa}_i$, correspond to those variables $\mathbf{Pa}_i$ that satisfy

$$p(x_i|\mathbf{pa}_i) = p(x_i|x_1, \ldots, x_{i-1}, x_{i+1}, \ldots, x_n) \quad (1)$$

The distributions in $\mathcal{P}$ are the local probability distributions $p(x_i|\mathbf{pa}_i), i = 1, \ldots, n$. We do not require the distributions $p(x_i|x_1, \ldots, x_{i-1}, x_{i+1}, \ldots, x_n), i = 1, \ldots, n$ to be obtainable (via inference) from a single joint distribution $p(\mathbf{x})$. If they are, we say that the dependency network is *consistent* with $p(\mathbf{x})$. We shall say more about the issue of consistency later in this section.

A Bayesian network for $\mathbf{X}$ defines a joint distribution for $\mathbf{X}$ via the product of its local distributions. A dependency network for $\mathbf{X}$ also defines a joint distribution for $\mathbf{X}$, but in a more complicated way via a Gibbs sampler (e.g., Gilks, Richardson, and Spiegelhalter, 1996). In this Gibbs sampler, we initialize each variable to some arbitrary value. We then repeatedly cycle through each variable $X_1, \ldots, X_n$, in this order, and resample each $X_i$ according to $p(x_i|x_1, \ldots, x_{i-1}, x_{i+1}, \ldots, x_n) = p(x_i|\mathbf{pa}_i)$. We call this procedure an *ordered Gibbs sampler*. As described by the following theorem (also proved in Hofmann, 2000), this ordered Gibbs sampler defines a joint distribution for $\mathbf{X}$.

**Theorem 1:** An ordered Gibbs sampler applied to a dependency network for $\mathbf{X}$, where each $X_i$ is discrete and each local distribution $p(x_i|\mathbf{pa}_i)$ is positive, has a unique stationary joint distribution for $\mathbf{X}$.

**Proof:** Let $\mathbf{x}^t$ be the sample of $\mathbf{x}$ after the $t^{th}$ iteration of the ordered Gibbs sampler. The sequence $\mathbf{x}^1, \mathbf{x}^2, \ldots$ can be viewed as samples drawn from a homogenous Markov chain with transition matrix $M$ having elements $M_{j|i} = p(\mathbf{x}^{t+1} = j|\mathbf{x}^t = i)$. (We use the terminology of Feller, 1957.) It is not difficult to see that $M$ is the product $M^1 \cdot \ldots \cdot M^n$, where $M^k$ is the "local" transition matrix describing the resampling of $X_k$ according to the local distribution $p(x_k|\mathbf{pa}_k)$. The positivity of local distributions guarantees the positivity of $M$, which in turn guarantees (1) the irreducibility of the Markov chain and (2) that all of the states are ergodic. Consequently, there exists a unique joint distribution that is stationary with respect to $M$. □

Because the Markov chain described in the proof is irreducible and ergodic, after a sufficient number of iterations, the samples in the chain will be drawn from the stationary distribution for $\mathbf{X}$. Consequently, these samples can be used to estimate this distribution.

Note that the Theorem holds for both consistent and inconsistent dependency networks. Furthermore, the restriction to discrete variables can be relaxed, but will not be discussed here. In the remainder of this paper, we assume all variables are discrete and each local distribution is positive.

In addition to determining a joint distribution, a dependency network for a given domain can be used to compute any conditional distribution of interest—that is, perform probabilistic inference. We discuss an algorithm for doing so, which uses Gibbs sampling, in Heckerman, Chickering, Meek, Rounthwaite, and Kadie (2000). That Gibbs sampling is used for inference may appear to be a disadvantage of dependency networks with respect to Bayesian networks. When we learn a Bayesian network from data, however, the resulting structures are typically complex and not amenable to exact inference. In such situations, Gibbs sampling (or even more complicated Monte-Carlo techniques) are used for inference in Bayesian networks, thus weakening this potential advantage.

In fact, when we have data and can learn a model for $\mathbf{X}$, dependency networks have an advantage over Bayesian networks. Namely, we can learn each local distribution in a dependency network independently, without regard to acyclicity constraints.

Bayesian networks have one clear advantage over dependency networks. In particular, dependency networks are not suitable for the representation of causal relationships. For example, if $X$ causes $Y$ (so that $X$ and $Y$ are dependent), the corresponding dependency network is $X \leftrightarrow Y$—that is, $X$ is a parent of $Y$ and vice versa. It follows that dependency networks are difficult to elicit directly from experts. Without an underlying causal interpretation, knowledge-based elicitation is cumbersome at best.

Another important observation about dependency networks is that, when we learn one from data as we have described—learning each local distribution independently—the model is likely to be inconsistent. (In an extreme case, where (1) the true joint distribu-tion lies in one of the possible models, (2) the model search procedure finds the true model, and (3) we have essentially an infinite amount of data, the learned model will be consistent.) A simple approach to avoid this difficulty is to learn a Bayesian network and apply inference to that network to construct the dependency network. This approach, however, will eliminate the advantage associated with learning dependency networks just described, is likely to be computationally inefficient, and may produce extremely complex local distributions. When ordered Gibbs sampling is applied to an inconsistent dependency network, it is important to note that the joint distribution so defined will depend on the order in which the Gibbs sampler visits the variables. For example, consider the inconsistent dependency network $X \leftarrow Y$. If we draw sample-pairs $(x, y)$—that is, $x$ and then $y$—then the resulting stationary distribution will have $X$ and $Y$ independent. In contrast, if we draw sample-pairs $(y, x)$, then the resulting stationary distribution may have $X$ and $Y$ dependent.

The fact that we obtain a joint distribution from any dependency network, consistent or not, is comforting. A more important question, however, is what distribution do we get? The following theorem, proved in Heckerman et al. (2000), provides a partial answer.

**Theorem 2:** If a dependency network for $\mathbf{X}$ is consistent with a positive distribution $p(\mathbf{x})$, then the stationary distribution defined in Theorem 1 is equal to $p(\mathbf{x})$.

When a dependency network is inconsistent, the situation is even more interesting. If we start with learned local distributions that are only slight perturbations (in some sense) of the true local distributions, will Gibbs sampling produce a joint distribution that is a slight perturbation of the true joint distribution? Hofmann (2000) argues that, for discrete dependency networks with positive local distributions, the answer to this question is yes when perturbations are measured with an L2 norm. In addition, Heckerman et al. (2000) show empirically using several real datasets that the joint distributions defined by a Bayesian network and dependency network, both learned from data, are similar.

We close this section with several facts about consistent dependency networks, proved in Heckerman et al. (2000). We say that a dependency network for $\mathbf{X}$ is *bi-directional* if $X_i$ is a parent of $X_j$ if and only if $X_j$ is a parent of $X_i$, for all $X_i$ and $X_j$ in $\mathbf{X}$. We say that a distribution $p(\mathbf{x})$ is *consistent* with a dependency network structure if there exists a consistent dependency network with that structure that defines $p(\mathbf{x})$.

**Theorem 3:** The set of positive distributions consistent with a dependency network structure is equal to the set of positive distributions defined by a Markov-network structure with the same adjacencies.

Note that, although dependency networks and Markov networks define the same set of distributions, their representations are quite different. In particular, the dependency network includes a collection of conditional distributions, whereas the Markov network includes a collection of joint potentials.

Let $\mathbf{pa}_i^j$ be the $j^{th}$ parent of node $X_i$. A consistent dependency network is *minimal* if and only if, for every node $X_i$ and for every parent $\mathbf{pa}_i^j$, $X_i$ is not independent of $\mathbf{pa}_i^j$ given the remaining parents of $X_i$.

**Theorem 4:** A minimal consistent dependency network for a positive distribution $p(\mathbf{x})$ must be bi-directional.

## 3 Learning Dependency Networks

In this section, we mention a few important points about learning dependency networks from data.

When learning a dependency network for $\mathbf{X}$, each local distribution for $X_i$ is simply a regression/classification model (with feature selection) for $x_i$ with $\mathbf{X} \setminus \{x_i\}$ as inputs. If we assume that each local distribution has a parametric model $p(x_i|\mathbf{pa}_i, \theta_i)$, and ignore the dependencies among the parameter sets $\theta_1, \ldots, \theta_n$, then we can learn each local distribution independently using any regression/classification technique for models such as a generalized linear model, a neural network, a support-vector machine, or an embedded regression/classification model (Heckerman and Meek, 1997). From this perspective, the dependency network can be thought of as a mechanism for combining regression/classification models via Gibbs sampling to determine a joint distribution.

In the work described in this paper, we use decision trees for the local distributions. A good discussion of methods for learning decision trees is given in Breiman, Friedman, Olshen, and Stone (1984). We learn a decision tree using a simple hill-climbing approach in conjunction with a Bayesian score as described in Friedman and Goldszmdit (1996) and Chickering, Heckerman, and Meek (1997). To learn a decision tree for $X_i$, we initialize the search algorithm with a singleton root node having no children. Then, we replace each leaf node in the tree with a binary split on some variable $X_j$ in $\mathbf{X} \setminus X_i$, until no such replacement increases the score of the tree. Our *binary split* on $X_j$ is a decision-tree node with two children: one of the children corresponds to a particular value of $X_j$, and the other child corresponds to *all other* values of $X_j$. Our Bayesian scoring function uses a uniform prior distribution for all decision-tree parameters, and a structure prior proportional to $\kappa^f$, where $\kappa > 0$ is a tunable parameter and $f$ is the number of free parameters in the decision tree. In studies that predated those described in this paper, we have found that the setting $\kappa = 0.01$ yields accurate models over a wide variety of datasets. We use this same setting in our experiments.

For comparison in these experiments, we also learn Bayesian networks with decision trees for local distributions using the algorithm described in Chickering, Heckerman, and Meek (1997). When learning these networks, we use the same parameter and structure priors used for dependency networks.

We conclude this section by noting an interesting fact about the decision-tree representation of local distributions. Namely, there will be a split on variable $X$ in the decision tree for $Y$ if and only if there is an arc from $X$ to $Y$ in the dependency network that includes these variables. As we shall see in Section 5, this correspondence helps the visualization of data.

## 4 Collaborative Filtering

In the remainder of this paper, we consider useful applications of dependency networks, whether they be consistent or not.

The first application is collaborative filtering (CF), the task of predicting preferences. Examples of this task include predicting what movies a person will like based on his or her ratings of movies seen, predicting what new stories a person is interested in based on other stories he or she has read, and predicting what web pages a person will go to next based on his or her history on the site. Another important application in the burgeoning area of e-commerce is predicting what products a person will buy based on products he or she has already purchased and/or dropped into his or her shopping basket.

Collaborative filtering was introduced by Resnick, Iacovou, Suchak, Bergstrom, and Riedl (1994) as both the task of predicting preferences and a class of algorithms for this task. The class of algorithms they described was based on the informal mechanisms people use to understand their own preferences. For example, when we want to find a good movie, we talk to other people that have similar tastes and ask them what they like that we haven't seen. The type of algorithm introduced by Resnik et al. (1994), sometimes called a *memory-based algorithm*, does something similar. Given a user's preferences on a series of items, the algorithm finds similar users in a database of stored preferences. It then returns some weighted average of preferences among these users on items not yet rated by the original user.

As done in Breese, Heckerman, and Kadie (1998), let us concentrate on the *application* of collaborative filtering—that is, preference prediction. In their paper, Breese et al. (1998) describe several CF scenarios, including binary versus non-binary preferences and implicit versus explicit voting. An example of explicit voting would be movie ratings provided by a user. An example of implicit voting would be knowing only whether a person has or has not purchased a product. Here, we concentrate on one scenario important for e-commerce: implicit voting with binary preferences—for example, the task of predicting what products a person will buy, knowing only what other products they have purchased.

A simple approach to this task, described in Breese et al. (1998), is as follows. For each item (e.g., product), define a variable with two states corresponding to whether or not that item was preferred (e.g., purchased). We shall use "0" and "1" to denote not preferred and preferred, respectively. Next, use the dataset of ratings to learn a Bayesian network for the joint distribution of these variables $\mathbf{X} = (X_1, \ldots, X_n)$. The preferences of each user constitutes a case in the learning procedure. Once the Bayesian network is constructed, make predictions as follows. Given a new user's preferences $\mathbf{x}$, use the Bayesian network to determine $p(X_i = 1|\mathbf{x} \setminus X_i = 0)$ for each product $X_i$ not purchased. That is, infer the probability that the user would have purchased the item had we not known he did not. Then, return a list of recommended products—among those that the user did not purchase—ranked by this probability.

Breese et al. (1998) show that this approach outperforms memory-based and cluster-based methods on several implicit rating datasets. Specifically, the Bayesian-network approach was more accurate and yielded faster predictions than did the other methods.

What is most interesting about this algorithm in the context of this paper is that only the probabilities $p(X_i = 1|\mathbf{x} \setminus X_i = 0)$ are needed to produce the recommendations. In particular, these probabilities may be obtained by a direct lookup in a dependency network:

$$p(X_i = 1|\mathbf{x} \setminus X_i = 0) = p(X_i = 1|\mathbf{pa}_i) \quad (2)$$

where $\mathbf{pa}_i$ is the instance of $\mathbf{Pa}_i$ consistent with $\mathbf{X}$. Thus, dependency networks are a natural model on which to base CF predictions. In the remainder of this section, we compare this approach with that based

Table 1: Number of users, items, and items per user for the datasets used in evaluating the algorithms.

| | Dataset | | |
|---|---|---|---|
| | MS.COM | Nielsen | MSNBC |
| Users in training set | 32,711 | 1,637 | 10,000 |
| Users in test set | 5,000 | 1,637 | 10,000 |
| Total items | 294 | 203 | 1,001 |
| Mean items per user in training set | 3.02 | 8.64 | 2.67 |

on Bayesian networks for datasets containing binary implicit ratings.

### 4.1 Datasets

We evaluated Bayesian networks and dependency networks on three datasets: (1) *Nielsen*, which records whether or not users watched five or more minutes of network TV shows aired during a two-week period in 1995 (made available courtesy of Nielsen Media Research), (2) *MS.COM*, which records whether or not users of microsoft.com on one day in 1996 visited areas ("vroots") of the site (available on the Irvine Data Mining Repository), and (3) *MSNBC*, which records whether or not visitors to MSNBC on one day in 1998 read stories among the most popular 1001 stories on the site. The MSNBC dataset contains 20,000 users sampled at random from the approximate 600,000 users that visited the site that day. In a separate analysis on this dataset, we found that the inclusion of additional users did not produce a substantial increase in accuracy. Table 4.1 provides additional information about each dataset. All datasets were partitioned into training and test sets at random.

### 4.2 Evaluation Criteria and Experimental Procedure

We have found the following three criteria for collaborative filtering to be important: (1) the accuracy of the recommendations, (2) prediction time—the time it takes to create a recommendation list given what is known about a user, and (3) the computational resources needed to build the prediction models. We measure each of these criteria in our empirical comparison. In the remainder of this section, we describe our evaluation criterion for accuracy.

Our criterion attempts to measure a user's expected utility for a list of recommendations. Of course, different users will have different utility functions. The measure we introduce provides what we believe to be a good approximation across many users.

The scenario we imagine is one where a user is shown a ranked list of items and then scans that list for preferred items starting from the top. At some point, the user will stop looking at more items. Let $p(k)$ denote the probability that a user will examine the $k$th item on a recommendation list before stopping his or her scan, where the first position is given by $k = 0$. Then, a reasonable criterion is

$$\text{cfaccuracy}_1(\text{list}) = \sum_k p(k)\ \delta_k \tag{3}$$

where $\delta_k$ is 1 if the item at position $k$ is preferred and 0 otherwise. To make this measure concrete, we assume that $p(k)$ is an exponentially decaying function:

$$p(k) = 2^{-k/a} \tag{4}$$

where $a$ is the "half-life" position—the position at which an item will be seen with probability 0.5. In our experiments, we use $a = 5$.

In one possible implementation of this approach, we could show recommendations to a series of users and ask them to rate them as "preferred" or "not preferred". We could then use the average of $\text{cfaccuarcy}_1(\text{list})$ over all users as our criterion. Because this method is extremely costly, we instead use an approach that uses only the data we have. In particular, as already described, we randomly partition a dataset into a training set and a test set. Each case in the test set is then processed as follows. First, we randomly partition the user's preferred items into *input* and *measurement* sets. The input set is fed to the CF model, which in turn outputs a list of recommendations. Finally, we compute our criterion as

$$\text{cfaccuracy}(\text{list}) = \frac{100}{N} \sum_{i=1}^{N} \frac{\sum_k \delta_{ik}\ p(k)}{\sum_{k=0}^{K_i-1} p(k)} \tag{5}$$

where $N$ is the number of users in the test set, $K_i$ is the number of preferred items in the measurement set for user $i$, and $\delta_{ik}$ is 1 if the $k$th item in the recommendation list for user $i$ is preferred in the measurement set and 0 otherwise. The denominator in Equation 5 is a per-user normalization factor. It is the utility of a list where all preferred items are at the top. This normalization allows us to more sensibly combine scores across measurement sets of different size.

We performed several experiments reflecting differing numbers of ratings available to the CF engines. In the first protocol, we included all but one of the preferred items in the input set. We term this protocol *all but 1.* In additional experiments, we placed 2, 5, and 10 preferred items in the input sets. We call these protocols *given 2*, *given 5*, and *given 10.*

The *all but 1* experiments measure the algorithms' performance when given as much data as possible from each test user. The various *given* experiments look at users with less data available, and examine the performance of the algorithms when there is relatively little known about an active user. When running the *given* $m$ protocols, if an input set for a given user had less than $m$ preferred items, the case was eliminated from the evaluation. Thus the number of trials evaluated under each protocol varied.

All experiments were performed on a 300 MHz Pentium II with 128 MB of memory, running the NT 4.0 operating system.

### 4.3 Results

Table 2 shows the accuracy of recommendations for Bayesian networks and dependency networks across the various protocols and three datasets. For a comparison, we also measured the accuracy of recommendation lists produced by sorting items on their overall popularity, $p(X_i = 1)$. The accuracy of this approach is shown in the row labeled "Baseline." A score in boldface corresponds to a significantly significant winner. We use ANOVA (e.g., McClave and Dietrich, 1988) with $\alpha = 0.1$ to test for statistical significance. When the difference between two scores in the same column exceed the value of RD (required difference), the difference is significant.

From the table, we see that Bayesian networks are more accurate than dependency networks. This result is interesting, because there are reasons to expect that dependency networks will be more accurate than Bayesian networks and vice versa. On the one hand, the search process that learns Bayesian networks is constrained by acyclicity, suggesting that dependency networks may be more accurate. On the other hand, the conditional probabilities used to sort the recommendations are *inferred* from the Bayesian network, but learned directly in the dependency network. Therefore, dependency networks may be less accurate, because they waste data in the process of learning what could otherwise be inferred. For this or perhaps other reasons, the Bayesian networks are more accurate.

The magnitudes of accuracy differences, however, are not that large. In particular, the ratio of (cfaccuracy(BN) - cfaccuracy(DN)) to (cfaccuracy(BN) - cfaccuracy(Baseline)) averages $4 \pm 5$ percent across the datasets and protocols.

Tables 3 and 4 compare the two methods with the remaining criteria. Here, dependency networks are a clear winner. They are significantly faster at prediction—sometimes by almost an order of magnitude—and require substantially less time and memory to learn.

Table 2: CF accuracy for the MS.COM, Nielsen, and MSNBC datasets. Higher scores indicate better performance. Statistically significant winners are shown in boldface.

| | MS.COM | | | |
|---|---|---|---|---|
| Algorithm | Given2 | Given5 | Given10 | AllBut1 |
| BN | **53.18** | 52.48 | 51.64 | 66.54 |
| DN | 52.68 | 52.54 | 51.48 | 66.60 |
| *RD* | *0.30* | *0.73* | *1.62* | *0.34* |
| Baseline | 43.37 | 39.34 | 39.32 | 49.77 |

| | Nielsen | | | |
|---|---|---|---|---|
| Algorithm | Given2 | Given5 | Given10 | AllBut1 |
| BN | **24.99** | 30.03 | 33.84 | **45.55** |
| DN | 24.20 | 29.71 | 33.80 | 44.30 |
| *RD* | *0.32* | *0.40* | *0.65* | *0.72* |
| Baseline | 12.65 | 12.72 | 12.92 | 13.59 |

| | MSNBC | | | |
|---|---|---|---|---|
| Algorithm | Given2 | Given5 | Given10 | AllBut1 |
| BN | **40.34** | **34.20** | 30.39 | **49.58** |
| DN | 38.84 | 32.53 | 30.03 | 48.05 |
| *RD* | *0.35* | *0.77* | *1.54* | *0.39* |
| Baseline | 28.73 | 20.58 | 14.93 | 32.94 |

Overall, Bayesian networks are slightly more accurate but much less attractive from a computational perspective.

## 5 Data Visualization

Bayesian networks are well known to be useful for visualizing causal relationships. In many circumstances, however, analysts are only interested in predictive—that is, dependency and independency—relationships. In our experience, the directed-arc semantics of Bayesian networks interfere with the visualization of such relationships.

As a simple example, consider the Bayesian network $X \to Y$. Those familiar with the semantics of Bayesian networks immediately recognize that observing $Y$ helps to predict $X$. Unfortunately, the untrained individual will not. In our experience, this person will interpret this network to mean that only $X$ helps to predict $Y$, and not vice versa. Even people who are expert in *d*-separation semantics will sometimes have difficulties visualizing predictive relationships using a Bayesian network. The cognitive act of identifying a node's Markov blanket seems to interfere with the visualization experience.

Dependency networks are a natural remedy to this problem. If there is no arc from $X$ to $Y$ in a dependency network, we know immediately that $X$ does not help to predict $Y$.

Figure 1 shows a dependency network learned from a dataset obtained from Media Metrix. The dataset contains demographic and internet-use data for about 5,000 individuals during the month of January 1997. On first inspection of this network, an interesting observation becomes apparent: there are many (predictive) dependencies among demographics, and many dependencies among frequency-of-use, but there are few dependencies between demographics and frequency-of-use.

Over the last three years, we have found numerous interesting dependency relationships across a wide variety of datasets using dependency networks for visualization. In fact, we have given dependency networks this name because they have been so useful in this regard.

The network in Figure 1 is displayed in DNViewer, a dependency-network visualization tool developed at Microsoft Research. The tool allows a user to display both the dependency-network structure and the decision tree associated with each variable. Navigation between the views is straightforward. To view a decision tree for a variable, a user simply double clicks on the corresponding node in the dependency network. Figure 2 shows the tree for Shopping.Freq.

An inconsistent dependency net learned from data offers an additional advantage for visualization. If there is an arc from $X$ to $Y$ in such a network, we know that $X$ is a *significant* predictor of $Y$—significant in whatever sense was used to learn the network. Under this interpretation, a uni-directional link between $X$ and $Y$ is not confusing, but rather informative. For example, in Figure 1, we see that Sex is a significant predictor of Socioeconomic status, but not vice versa—an interesting observation. Of course, when making such interpretations, one must always be careful to recognize that statements of the form "$X$ helps to predict $Y$" are made in the context of the other variables in the network.

In DNViewer, we enhance the ability of dependency networks to reflect strength of dependency by including a slider (on the left). As a user moves the slider from bottom to top, arcs are added to the graph in the order in which arcs are added to the dependency network during the learning process. When the slider is in its upper-most position, all arcs (i.e., all significant dependencies) are shown.

Table 3: Number of predictions per second for the MS.COM, Nielsen, and MSNBC datasets.

| | **MS.COM** | | | |
|---|---|---|---|---|
| Algorithm | Given2 | Given5 | Given10 | AllBut1 |
| BN | 3.94 | 3.84 | 3.29 | 3.93 |
| DN | 23.29 | 19.91 | 10.20 | 23.48 |

| | **Nielsen** | | | |
|---|---|---|---|---|
| Algorithm | Given2 | Given5 | Given10 | AllBut1 |
| BN | 22.84 | 21.86 | 20.83 | 23.53 |
| DN | 36.17 | 36.72 | 34.21 | 37.41 |

| | **MSNBC** | | | |
|---|---|---|---|---|
| Algorithm | Given2 | Given5 | Given10 | AllBut1 |
| BN | 7.21 | 6.96 | 6.09 | 7.07 |
| DN | 11.88 | 11.03 | 8.52 | 11.80 |

Table 4: Computational resources for model learning.

| | **MS.COM** | |
|---|---|---|
| Algorithm | Memory (Meg) | Learn Time (sec) |
| BN | 42.4 | 144.65 |
| DN | 5.3 | 98.31 |

| | **Nielsen** | |
|---|---|---|
| Algorithm | Memory (Meg) | Learn Time (sec) |
| BN | 3.3 | 7.66 |
| DN | 2.1 | 6.47 |

| | **MSNBC** | |
|---|---|---|
| Algorithm | Memory (Meg) | Learn Time (sec) |
| BN | 43.0 | 105.76 |
| DN | 3.7 | 96.89 |

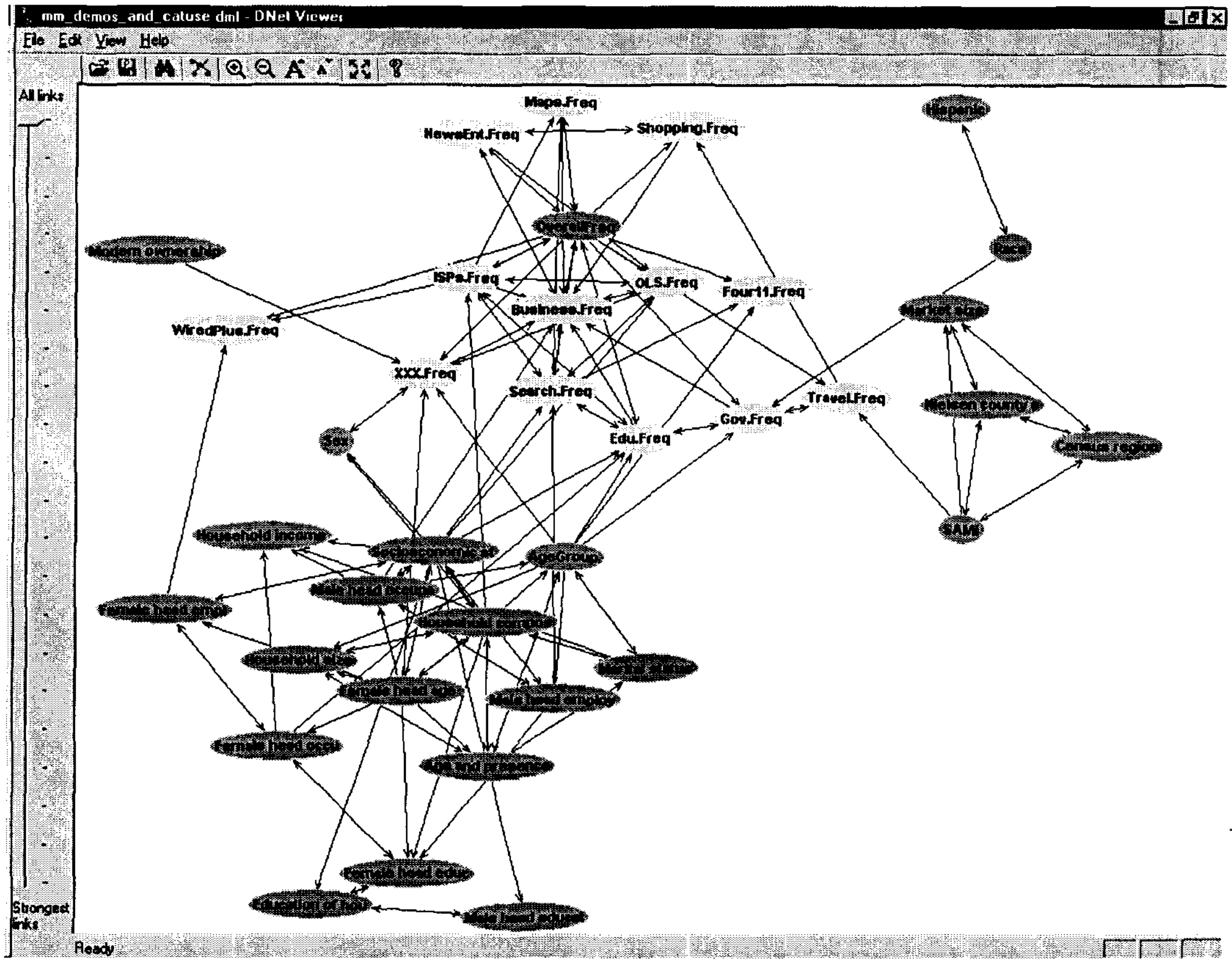

Figure 1: A dependency network for Media Metrix data. The dataset contains demographic and internet-use data for about 5,000 individuals during the month of January 1997. The node labeled Overall.Freq represents the overall frequency-of-use of the internet during this period. The nodes Search.Freq, Edu.Freq, and so on represent frequency-of-use for various subsets of the internet.

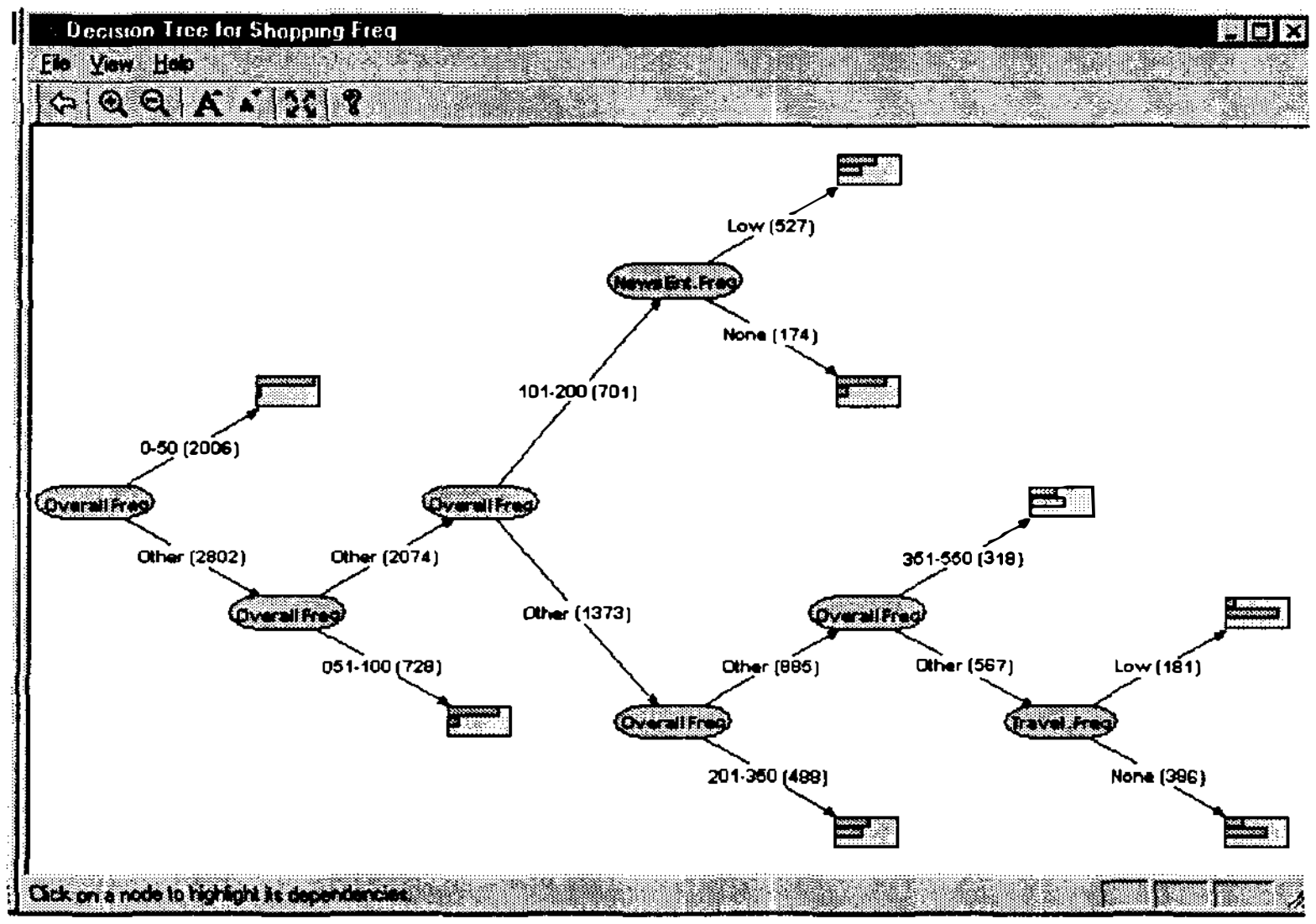

Figure 2: The decision tree for Shopping.Freq obtained by double-clicking that node in the dependency network. The histograms at the leaves correspond to probabilities of Shopping.Freq use being zero, one, and greater than one visit per month, respectively.

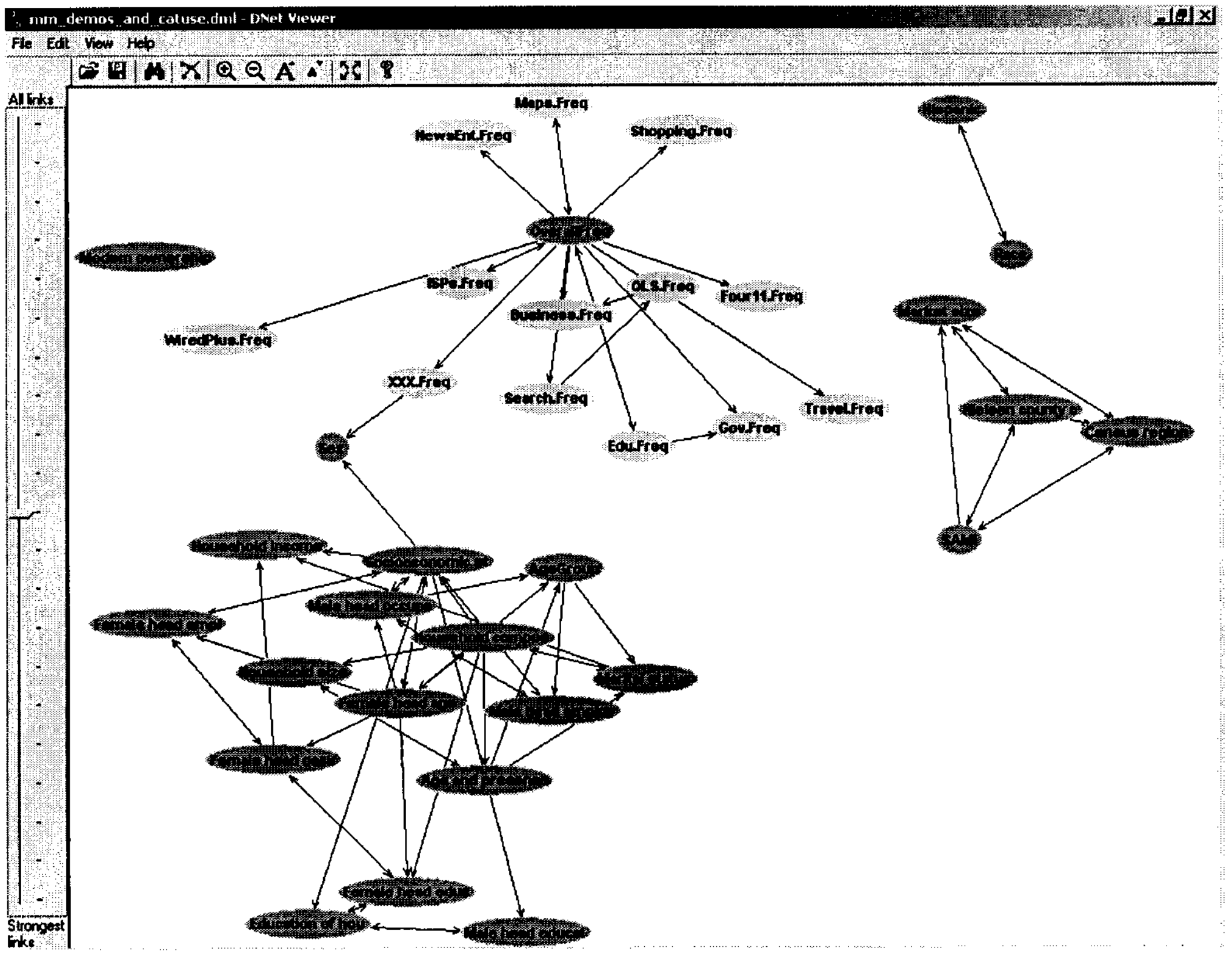

Figure 3: The dependency network in Figure 1 with the slider set at half position.

Figure 3 shows the dependency network for the Media Metrix data with the slider at half position. At this setting, we find the interesting observation that the dependencies between Sex and XXX.Freq (frequency of hits to pornographic pages) are the strongest among all dependencies between demographics and internet use.

## 6 Summary and Future Work

We have described a new representation for probabilistic dependencies called a dependency network. We have shown that a dependency network (consistent or not) defines a joint distribution for its variables, and that models in this class are easy to learn from data. In particular, we have shown how a dependency network can be thought of as a collection of regression/classification models among variables in a domain that can be combined using Gibbs sampling to define a joint distribution for the domain. In addition, we have shown that this representation is useful for collaborative filtering and the visualization of predictive relationships.

Of course, this research is far from complete. There are many questions left to be answered. For example, what are useful models (e.g., generalized linear models, neural networks, support-vector machines, or embedded regression/classification models) for a dependency network's local distributions? Another example of particular theoretical interest is Hofmann's (2000) result that small L2-norm perturbations in the local distributions lead to small L2-norm perturbations in the joint distributions defined by the dependency network. Can this result be extended to norms more appropriate for probabilities such as cross entropy?

Finally, the dependency network and Bayesian network can be viewed as two extremes of a spectrum. The dependency network is ideal for situations where the conditionals $p(x_i|\mathbf{x} \setminus x_i)$ are needed. In contrast, when we require the joint probabilities $p(\mathbf{x})$, the Bayesian network is ideal because these probabilities may be obtained simply by multiplying conditional probabilities found in the local distributions of the variables. In situations where we need probabilities of the form $p(\mathbf{y}|\mathbf{x} \setminus \mathbf{y})$, where $\mathbf{Y}$ is a proper subset of the domain $\mathbf{X}$, we can build a network structure that enforces an acyclicity constraint among only the variables $\mathbf{Y}$. In so doing, the conditional probabilities $p(\mathbf{y}|\mathbf{x} \setminus \mathbf{y})$ can be obtained by multiplication.

## Acknowledgments

We thank Reimar Hofmann for useful discussions. Datasets for this paper were generously provided by Media Metrix, Nielsen Media Research (Nielsen), Microsoft Corporation (MS.COM), and Steven White and Microsoft Corporation (MSNBC).